\documentclass[letterpaper, 10pt, conference]{ieeeconf}
\IEEEoverridecommandlockouts
\usepackage{tabularx} % needs to be first
\usepackage[utf8]{inputenc}
\usepackage[T1]{fontenc}
\usepackage{acro}
\usepackage{amsmath}
\usepackage{amssymb}
\usepackage{amsfonts}
\usepackage{arydshln}
\usepackage{balance}
\usepackage{bm} % bold math with \bm
\usepackage{booktabs}
\usepackage[font=small]{caption}
\usepackage{graphicx}
\usepackage{multirow}
\usepackage{pgfplots}
\usepackage{pgfplotstable}
\usepackage[separate-uncertainty=true, range-units=single]{siunitx}
\usepackage{tikz}
\usepackage{url}
\usepackage{xcolor}

\let\labelindent\relax % is already defined in IEEE template
\usepackage{enumitem}

\usetikzlibrary{backgrounds, calc, positioning, 3d, scopes, shapes.misc, math}
\pgfplotsset{compat=1.17}
\usepgfplotslibrary{fillbetween}

\DeclareAcronym{NN}{short=NN, long=neural network}
\DeclareAcronym{DoF}{short=DoF, long=degree of freedom, long-plural-form=degrees of freedom}

\renewcommand{\vec}[1]{\bm{#1}}
\newcommand{\sysfull}{SpeedFolding}
\newcommand{\algfull}{BiManual Manipulation Network}
\newcommand{\algname}{BiMaMa-Net}
\newcommand{\done}{Sufficiently Smoothed}

\title{\LARGE \bf
SpeedFolding: Learning Efficient Bimanual Folding of Garments
}

\author{
    Yahav Avigal$^{*,1}$, Lars Berscheid$^{*,1,2}$, Tamim Asfour$^{2}$, Torsten Kröger$^{2}$, and Ken Goldberg$^{1}$
    \thanks{$^{1}$AUTOLab at UC Berkeley
	    {\tt\small \{yahav\_avigal, goldberg\}@berkeley.edu}
	}
    \thanks{$^{2}$Karlsruhe Institute of Technology (KIT)
	    {\tt\small \{lars.berscheid, asfour, torsten\}@kit.edu}
	}
	\thanks{$^{*}$Equal contribution: order determined by shirt folding skills.
	}
}

\IEEEoverridecommandlockouts

\begin{document}

\maketitle

\thispagestyle{empty}
\pagestyle{empty}

%%%%%%%%%%%%%%%%%%%%%%%%%%%%%%%%%%%%%%%%%%%%%%%%%%%%%%%%%%%%%%%%%%%%%%%%%%%%%%%%
\begin{abstract}

Folding garments reliably and efficiently is a long standing challenge in robotic manipulation due to the complex dynamics and high dimensional configuration space of garments. An intuitive approach is to initially manipulate the garment to a canonical smooth configuration before folding. In this work, we develop SpeedFolding, a reliable and efficient bimanual system, which given user-defined instructions as folding lines, manipulates an initially crumpled garment to (1) a smoothed and (2) a folded configuration.
Our primary contribution is a novel neural network architecture that is able to predict pairs of gripper poses to parameterize a diverse set of \emph{bimanual} action primitives.
After learning from \num{4300} human-annotated and self-supervised actions, the robot is able to fold garments from a random initial configuration in under \SI{120}{s} on average with a success rate of \SI{93}{\%}.
Real-world experiments show that the system is able to generalize to unseen garments of different color, shape, and stiffness. While prior work achieved 3-6 Folds Per Hour (FPH), SpeedFolding achieves 30-40 FPH.
% SpeedFolding decreases the folding time by over \SI{30}{\%} in comparison to baselines, and outperforms prior works requiring \SIrange{10}{20}{min} per fold by \SIrange{5}{10}{\times}.

See \url{https://pantor.github.io/speedfolding} for code, videos, and datasets.

% SpeedFolding can manipulate garments to a \textit{Ready to Fold} configuration and then follow user-defined instructions to generate a desired fold, or leverage prior information on the garment's dimensions to fold it directly from an intermediate configuration.
% We compare SpeedFolding to baseline garment-smoothing and folding methods in real-world experiments. Results suggest that SpeedFolding can fold initially crumpled garments reliably, including garments unseen during training with different color, shape and stiffness, decreasing folding time from 30 -- 47\,\%. 
% While prior works have folded a t-shirt in 10-20 minutes, SpeedFolding can fold a t-shirt in under 2 minutes, and given prior knowledge of the t-shirt it can fold in under 90 seconds.

\end{abstract}

%%%%%%%%%%%%%%%%%%%%%%%%%%%%%%%%%%%%%%%%%%%%%%%%%%%%%%%%%%%%%%%%%%%%%%%%%%%%%%%%
\section{Introduction}
\label{sec:introduction}

Garment handling such as folding and packing are common tasks in textile manufacturing and logistics, industrial and household laundry, healthcare, and hospitality, where speed and efficiency are key factors. These tasks are largely performed by humans due to the complex configuration space as well as the highly non-linear dynamics of deformable objects~\cite{zhu2021challenges, ganapathi2021learning}. 
Additionally, folding is a long horizon sequential planning problem, as it requires to first flatten or smooth the garment, and then follow a sequence of steps~\cite{doumanoglou2016folding, maitin2010cloth} or sub-goals~\cite{weng2022fabricflownet} to achieve the desired fold. 

Prior work has mainly focused on single-arm manipulation~\cite{ganapathi2021learning,hoque2020visuospatial,seita2020deep, yunliang2022efficiently} or on complex iterative algorithms~\cite{doumanoglou2016folding, maitin2010cloth, bersch2011bimanual}, requiring a large number of interactions and resulting in long execution times. Recently, Ha et al.~\cite{ha2022flingbot} proposed a method for smoothing cloth that computes the pick points for a high-velocity dynamic fling action directly from overhead images, and can smooth garments to $80\,\%$ coverage in 3 actions on average. However, the proposed 4\,\acp{DoF} action parameterization constrains the two pick poses significantly, in particular by discrete distances and a fixed rotation in between.

We present \sysfull{}, an end-to-end system for fast and efficient garment folding. At first, a novel \algfull{} (\algname{}) learns to predict a pair of gripper poses for bimanual actions from an overhead RGBD input image to smooth an initially crumpled garment. Once the garment has been smoothed to a desired level, determined by a learned smoothing classifier, \sysfull{} executes a folding pipeline (see Fig.~\ref{fig:fig_one}). This paper contributes:
\input{figures/figure_one}
\input{figures/primitives}
% \vspace{-10pt}
\begin{enumerate}
    \item The \algname{} architecture for bimanual manipulation that computes two \emph{corresponding} planar gripper poses without any spatial restrictions, with an automated calibration procedure to account for robot reachability constraints.
    \item An end-to-end robotic system for efficient smoothing and folding. First, the system learns to smooth a garment to a sufficiently smoothed configuration through self-supervision. Then, the robot folds the garment according to user-defined folding lines.
    \item An experimental dataset from physical experiments that suggests the system can fold garments with a success rate of over \SI{90}{\%}, including garments unseen during training that differ in color, shape and stiffness. Folding a t-shirt takes under \SI{120}{s} on average, improving baselines by \SIrange{30}{47}{\%} and prior works by \SIrange{5}{10}{\times}.
\end{enumerate}

\section{Related Work}

\textbf{Bimanual robotic manipulation} has been studied extensively in fields from surgical robotics to industrial manipulation~\cite{smith2012dual}. A dual-arm system extends the workspace, allows for increased payload and for more complex behaviours than a single arm system~\cite{weng2022fabricflownet,lertkultanon2018certified, hayakawa2021dual, hu2018robotic}, but comes at the cost of higher planning complexity due to the additional \acp{DoF} and self-collisions~\cite{edsinger2007two}. A promising line of research is to employ dual-arm systems for garment manipulation~\cite{garcia2020benchmarking}. Garments are especially difficult to control and manipulate due to their large configuration space, self-occlusions, and complex dynamics~\cite{ganapathi2021learning}. Recent works have mainly focused on garment smoothing from arbitrary configurations~\cite{ha2022flingbot}, or garment folding, assuming the garment has been initially flattened~\cite{weng2022fabricflownet}. We present an end-to-end approach to smoothing and then folding garments from initial crumpled configurations.

\textbf{Garment smoothing} aims to transform the garment from an arbitrary crumpled configuration to a smooth configuration~\cite{seita2020deep}. Prior works have focused on extracting and identifying specific features such as corners and wrinkles~\cite{doumanoglou2016folding, maitin2010cloth, sun2013heuristic, willimon2011model}. Recent methods have used expert demonstrations to learn garment smoothing policies in simulation~\cite{ganapathi2021learning, hoque2020visuospatial, seita2020deep}, however these methods learn quasi-static pick-and-place actions that require a large number of interactions on initially crumpled garments. Ha et al.~\cite{ha2022flingbot} introduced a novel 4\,\ac{DoF} dynamic fling action parameterization learned in simulation that can achieve $\sim\SI{80}{\%}$ garment coverage within 3 actions. However, this parameterization is (1) limited to fling actions, (2) fails to fully smooth garments, and (3) induces grasp failures in more than \SI{25}{\%} of actions. In this work we use expert demonstrations and self-supervised learning purely in the physical world to train a novel bimanual manipulation \ac{NN} architecture to smooth a garment such that it is ready to be folded.

\textbf{Garment folding} has many applications in hospitals, homes and warehouses. Early approaches rely heavily on heuristics and can achieve high success rates, but have long cycle times on the order of \SIrange{10}{20}{min} per garment~\cite{doumanoglou2016folding, maitin2010cloth, bersch2011bimanual, balkcom2008robotic, tanaka2007origami}. Recent methods have been focusing on learning goal-conditioned policies in simulation~\cite{weng2022fabricflownet, hoque2020visuospatial, seita2021learning, tanaka2018emd} and directly on a physical robot~\cite{lee2020learning}. In this work, we compare an instruction-based folding approach that can reliably fold smoothed garments, with a novel folding approach that can fold a t-shirt directly from a non-smooth configuration given prior knowledge about its dimensions.  

\section{Problem Statement}
\label{sec:problem_statement}

\input{figures/bimamanet}

Given a visual observation $o_t \in \mathbb{R}^{W \times H \times C}$ of the garment's configuration $s_t$ at time $t$, the objective is to compute and execute an action $a_t$ to transfer the garment from an arbitrary configuration to a desired user-defined $s^*$ goal configuration. In particular, $s^*$ is invariant under the garment's position and orientation in the workspace.
We assume an overhead observation with a calibrated pixel-to-world transformation, as well as a garment that is easily distinguishable from the workspace.

We consider a dual-arm robot with parallel-jaw grippers executing actions of type $m \in \mathcal{M}$ from a discrete set of pre-defined action primitives. In particular, we parameterize each primitive by two planar gripper poses
\begin{align*}
    a_t &= \langle m, (x_1, y_1, \theta_1), (x_2, y_2, \theta_2) \rangle
\end{align*}
for each arm respectively, in which $(x_i, y_i)$ are coordinates in pixel space, and $\theta_i$ is the end-effector rotation about the $z$ axis. We further assume a flat obstacle-free workspace and a motion planner that computes collision-free trajectories for a dual-arm robot. 

\section{Method}
\label{sec:method}

\sysfull{} uses \algname{}, a learned garment-smoothing method to bring an initially crumpled garment to a sufficiently smooth configuration, followed by an instruction-based garment folding pipeline.

\subsection{Action Primitives}

We are interested in the set of quasi-static and dynamic action primitives that enable the robot to (1) transfer an arbitrary garment configuration $s_t$ to a folded goal configuration $s^*$ (completeness), (2) reducing the number of action steps (efficiency), and (3) with a reduced number of primitives (minimality). Each action primitive is defined through a pair of poses as well as a motion trajectory. All primitives share a common procedure to reliably grasp the garment with parallel jaw grippers: Each gripper moves $\SI{4}{cm}$ above the grasp pose $a_t$, rotates $8^\circ$ so that one fingertip is below the other, and moves $\SI{1}{cm}$ towards the direction of the higher fingertip. This motion improves the success for grasping in particular at the edge of the garment. We define following learned primitives (Fig.~\ref{fig:primitives} left box):

\begin{description}[leftmargin=4mm]
\item[Fling:] Given two pick poses, the arms first pick those points, lift the garment above the workspace and stretch it until a force threshold is reached, measured using the arms' internal force sensors. Next, the arms apply a dynamic motion, flinging the garment forward and then backward while gradually reducing the height toward the workspace. Similar to \cite{ha2022flingbot}, we find the fling motion to be robust under change of velocity and trajectory parameters, and therefore we keep these parameters fixed. The fling primitive allows to significantly increase the garment's coverage in a few steps, but often does not yield a smooth configuration.

\item[Pick-and-place:] Given a pick and a corresponding place pose, a single arm executes this quasi-static action, while the second arm presses down the garment at a point on a line extending the pick from the place pose. Pick-and-place enables the robot to fix local faults such as corners or sleeves folded on top of the garment.

\item[Drag:] Given two pick points, the robot drags the garment for a fixed distance away from the garment's center of mask, leveraging the friction with the workspace to smooth wrinkles or corners, e.g. sleeves folded below the garment.
\end{description}

We define heuristic-based primitives (Fig.~\ref{fig:primitives} center box):
\begin{description}[leftmargin=4mm]
\item[Fold:] Both arms execute a pick-and-place action simultaneously to fold the garment. The heuristic for calculating the pick and place poses is explained in Sec.~\ref{subsec:folding}.
\item[Move:] While similar to drag, this primitive's pick poses and its drag distance are calculated by a heuristic so that the garment's center of mask is moved to a goal target point. Usually, the robot drags the garment towards itself to mitigate reachability issues in subsequent actions. Sec.~\ref{subsec:folding} provides details about the pose calculation. 
\end{description}

We define an additional learned primitive to switch from garment-smoothing to folding (Fig.~\ref{fig:primitives} right box):

\begin{description}[leftmargin=4mm]
\item[\done{}:] We find that deciding whether a garment is ready to be folded purely from coverage computation, as done in prior works~\cite{hoque2020visuospatial, seita2020deep, ha2022flingbot}, is not sufficient.
In particular, even a high coverage is prone to wrinkles or faults that might reduce the subsequent fold quality significantly (as described in Fig.~\ref{fig:primitives}). Instead of relying on the coverage, \algname{} returns a smoothness value given an overhead image. While this primitive is not used to change the configuration of the garment, it is used to switch from garment-smoothing to folding.
\end{description}

\subsection{\algname{} for Bimanual Manipulation}
\label{sec:bimama}

Predicting a single pose from an overhead image is commonly done by first estimating a pixel-wise value map per gripper z-axis rotation $\theta$, in which each pixel value represents a future expected reward (e.g., grasp success, increase in garment coverage, etc.), and then selecting the maximum greedily \cite{zeng2018learning, berscheid2019robot, grannen2021untangling, ha2022flingbot}. Extending this approach to two \emph{corresponding} planar poses $(x, y, \theta)_{1,2}$ conditioned on each other is however challenging primarily due to the exponential scaling of possible end-effector poses with the number of dimensions. In particular, this is a multi-modal problem, and the predicted unconditioned value maps $Q_{unc}(x, y, \theta)$ have multiple peaks (as in Fig.~\ref{fig:reachability}). While unconditioned value maps may provide information relevant for downstream bimanual tasks, such as the grasp success, they provide no information regarding their correspondences. To address this we define \emph{correspondence descriptors}
\begin{align*}
    \vec{d} &= \left( Q_{unc}, x, y, \sin \theta, \cos \theta, m, \vec{e} \right)
\end{align*}
where $\vec{e} \in \mathbb{R}^M$ is a learned embedding for each pixel (disregarding orientations $\theta$) concatenated with the unconditioned value $Q_{unc}$, positional encodings, and the action primitive type $m$. Then, the final conditioned value $Q(\vec{d}_1, \vec{d}_2)$ depends on a descriptor pair.

\begin{figure}
    \centering
    \includegraphics{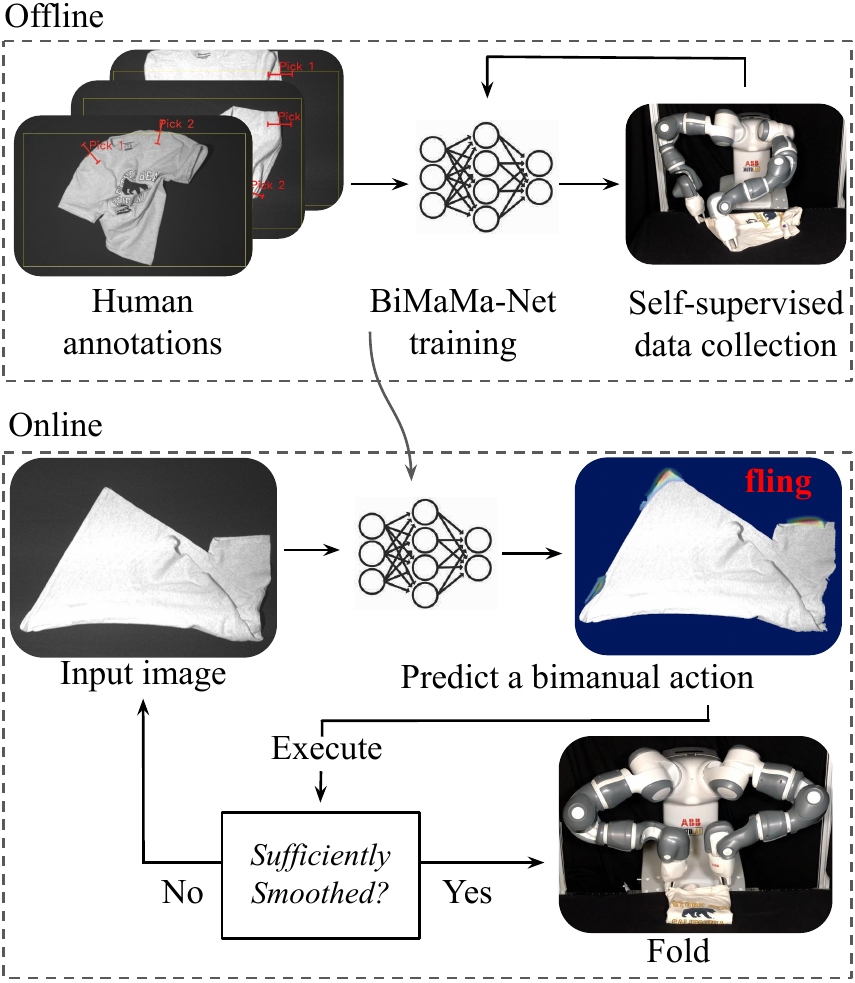}
    \caption{\textbf{SpeedFolding Pipeline.} We start by manually annotating input images with primitives and gripper poses, train a NN and then iteratively use the NN for self-supervised data collection (\textbf{top}). During runtime, we use the NN to predict a primitive and a pair of poses given an input image and execute it on the robot. If the resulting garment configuration is classified as \done{} the robot will fold the garment, otherwise it will repeat the process.}
    \label{fig:pipeline}
\end{figure}

Fig.~\ref{fig:bimamanet-architecture} shows the complete information flow of \algname{}: A shared encoder using a ResNext-50 \cite{xie2017aggregated} backbone maps an input image (e.g. depth and grayscale) to high-level features. First, a classification head predicts the manipulation primitive $m$. For a \emph{\done{}} primitive, no further action is required. For all other learned primitives, a U-Net~\cite{ronneberger2015u} decoder predicts value maps for a discrete number $N$ of end-effector orientations $\theta$. We choose a U-Net architecture over fully convolutional \ac{NN} used in prior robotics manipulation works~\cite{berscheid2020self, zeng2021transporter, ha2022flingbot, berscheid2019robot} as U-Nets are better suited for high-resolution inputs that we find necessary for detecting edges and wrinkles for garment smoothing.

Then, \emph{\algname{}} samples a set of poses from the value map, where pixels with higher values are more likely to get sampled. During training, \algname{} samples from $p(a|s) \sim \sqrt{Q_{unc}(a, s)}$ to allow for sampling negative examples to better estimate the underlying distribution of action values. For inference, \algname{} samples from $p(a|s) \sim Q_{unc}(a, s)^2$ which emphasizes action poses with high values. It then calculates the correspondence descriptors for each pose, and a final \ac{NN} head combines all descriptor pairs $\vec{d}_1$, $\vec{d}_2$ to output the final conditioned action value $Q$.

If two poses are interchangeable (e.g., during a fling or a drag action), a single decoder predicts the value maps per $\theta$. However, if a certain relation between the poses must be maintained (e.g., the conceptual difference between the pick and the place poses in a pick-and-place action) then separate decoders compute two value maps $Q_{unc}^1$ and $Q_{unc}^2$.

\subsection{Reachability Calibration}
\input{figures/folding}

\begin{figure}[b]
    \centering
\begin{tikzpicture}
    \node[] (r1) {\includegraphics[width=0.41\linewidth, trim=90 60 140 0, clip]{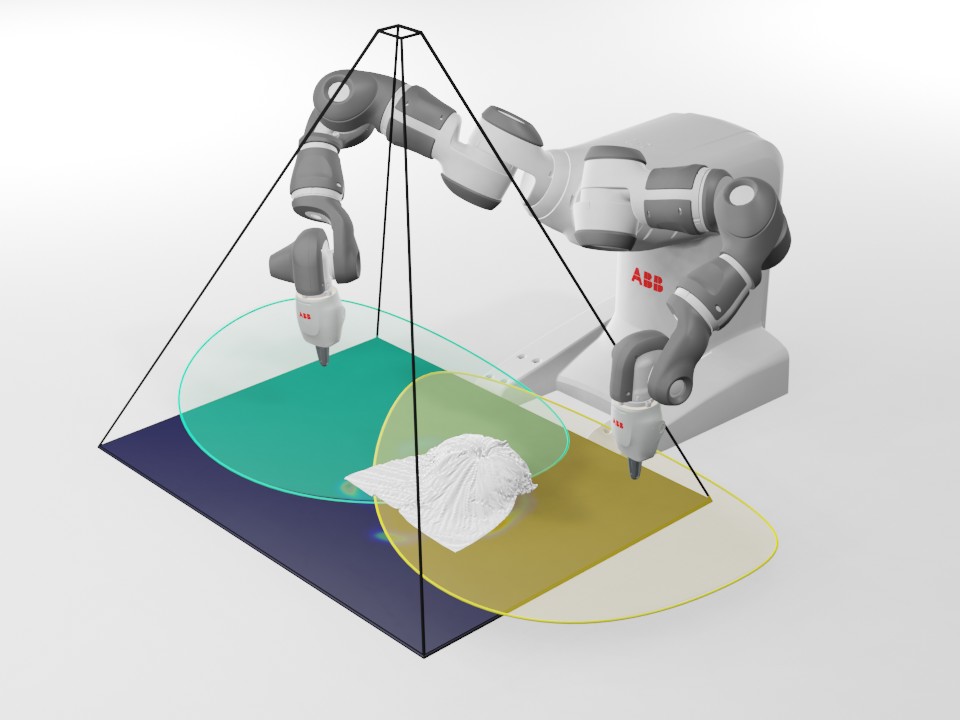}};
    \node[right=0mm of r1] (r2) {\includegraphics[width=0.51\linewidth, trim=12 0 0 0, clip]{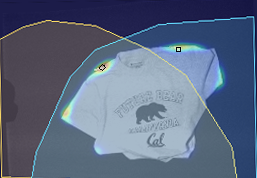}};
    
    \node[anchor=north, below=-1mm of r1, text width=112, align=center] {(a) Workspace};
    \node[anchor=north, below=-1mm of r2, text width=112, align=center] {(b) Reachability mask};
\end{tikzpicture}
    \caption{\textbf{Reachability.} (a) We perform a boundary search to compute separate reachability masks for the left (yellow) and right (blue) robot arms. (b) \algname{} guarantees at least one pick pose (black) from the value map within each mask.}
    \label{fig:reachability}
\end{figure}

As shown in Fig.~\ref{fig:reachability}, to ensure reliable garment smoothing and folding, the robot should compute the actions that maximize the expected reward within the reachable space. To find the robot's reachable space, we perform a one-time boundary search along a discretized grid in the action space $(x,y,\theta)$ for each gripper, assuming a constant height $z$ above the table. The search, done separately for each $\theta$, starts with a fixed lower value of $y$ and increases $x$ until the inverse kinematics fails to find a solution. Afterwards, it repeatedly increases $y$ or decreases $x$ so that the search is confined to the continuous boundary at which reachability fails. As a result, we get masks $M_l$ and $M_r$ for the left and right arms
\begin{align*}
    M(x, y, \theta) \rightarrow \lbrace 0, 1 \rbrace
\end{align*}
that can be incorporated into \algname{} as spatial binary constraints by restricting the action sampling to the masks. To ensure that each reachability mask contains at least one pose, 
we create up to four masked value maps from $Q_{unc}^{1}$ (or $Q_{unc}^{2}$) by multiplying them with $M_l$ or $M_r$: $\lbrace Q_{unc}^{1l}, Q_{unc}^{1r}, Q_{unc}^{2l}, Q_{unc}^{2r} \rbrace$. An action value $Q_{unc} = 0$ is ignored in the sampling process. We then sample and combine the correspondence descriptors from $Q_{unc}^{1l}$ and $Q_{unc}^{2r}$, and vice versa for $Q_{unc}^{1r}$ and $Q_{unc}^{2l}$.

We find that using a calibrated reachability mask for each end-effector orientation $\theta$ separately significantly reduces the number of false negatives that arise when using approximations, such as a circular mask. After selecting the final, reachable poses $(x, y, \theta)_1$ and $(x, y, \theta)_2$ during runtime, we check for possible collisions due to inter-arm interaction. If a potential collision is detected, the next best action is selected until reachable and collision-free poses are found.

\subsection{Training for Smoothing}
\label{sec:training}

We train \algname{} via
self-supervised real-world learning to predict the manipulation primitive type $m$ and the corresponding action poses $(x, y, \theta)_1$ and $(x, y, \theta)_2$ given an overhead image of a garment.

In order to scale real-world interaction, the learning process is designed for minimal human intervention. First, we collect examples of smooth configurations to train a classifier outputting the confidence $p(\textit{\done{}} | s)$. Additionally, let $\textit{cov}(s)$ be the coverage of the garment at configuration $s$ observed from an overhead perspective, calculated by background subtraction and color filtering.
We define the reward $r$:
\begin{align*}
    r_t &= \max\left( \tanh\left[ \alpha\, \left( \textit{cov}(s_{t+1}) - \textit{cov}(s_{t}) \right) \right.\right. \\
    &+ \left.\left. \, \beta\, \left( p(\textit{smoothed} | s_{t+1}) - p(\textit{smoothed} | s_{t}) \right) \right], 0\right)
\end{align*}
as the sum of the \emph{change} of coverage and \emph{\done{}} confidence with tuned weights $\alpha$ and $\beta$ respectively. It is scaled to $r \in [-1, 1]$ first and then clipped to a non-negative value, so that no change equals zero reward. To ensure continuous training, the robot resets the garment configuration by grasping it at a random position on its mask and dropping it from a fixed height. We iteratively train a self-supervised data collection \ac{NN}, interleaving training and execution (Fig.~\ref{fig:pipeline}). The robot explores different actions by uniformly sampling from the set of $N_s$ best actions.

To avoid a purely random and sample-inefficient initial exploration, we kickstart the training with human annotations. We differentiate between self-supervised and human annotated data within the training process in three ways: (1) We set the reward of human annotated data to a fixed $r_h$. (2) Besides training the value map at the specific annotated pixel position and orientation, we follow~\cite{grannen2021untangling} and introduce a Gaussian decay centered around each pose as a global target value instead.
(3) The classification head is trained only with data that has a reward higher than a tuned threshold $r \geq r_c$.

\subsection{Folding Pipeline}
\label{subsec:folding}
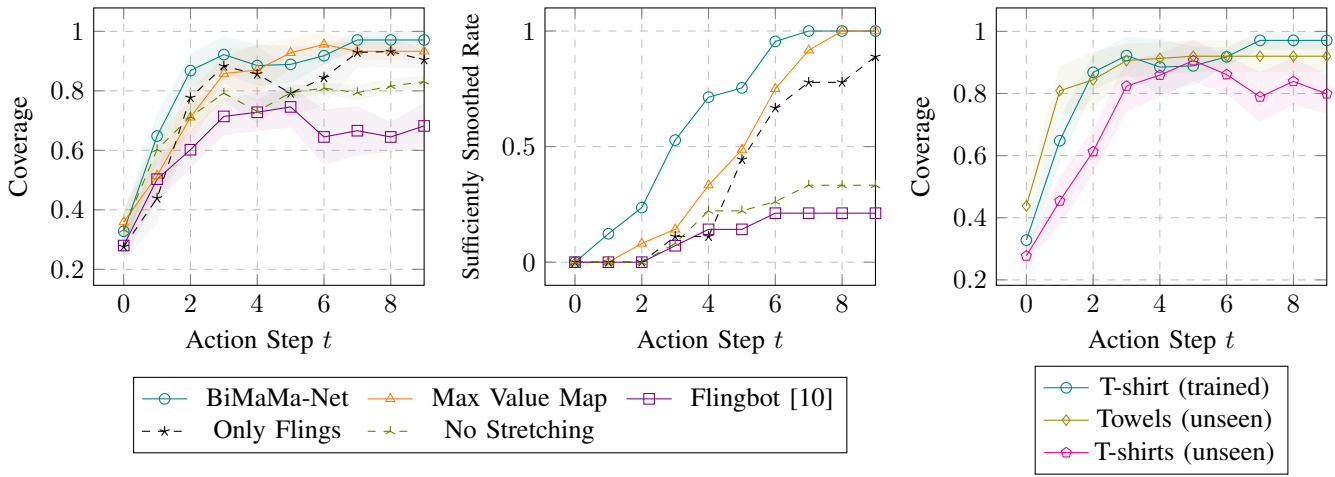
\begin{figure*}
    \vspace{1mm}
    \centering
\begin{tikzpicture}
    \tikzstyle{error} = [fill opacity=0.06]
    
\begin{axis}[
    xlabel={Action Step $t$},
    ylabel={Coverage},
    height=150,
    width=170,
    xmax=9,
    grid=both,
    grid style={dashed, draw=gray},
    major grid style={line width=.2pt,draw=gray!50},
    legend pos=south east,
    legend style={
        at={(1.20, -0.33)},
        anchor=north,
        legend columns=3,
        column sep=5pt
    }
]
\addplot+ [teal, mark=o] table[x=step, y=cov, teal] {figures/cov-over-step.txt};
\addlegendentry{BiMaMa-Net}

\addplot+ [orange, mark=triangle] table[x=step, y=cov_max] 
{figures/cov-over-step.txt};
\addlegendentry{Max Value Map}

\addplot+ [black, violet, mark=square] table[x=step, y=cov_flingbot] {figures/cov-over-step.txt};
\addlegendentry{Flingbot \cite{ha2022flingbot}}

\addplot+ [black, dashed, mark=star] table[x=step, y=cov_fling] {figures/cov-over-step.txt};
\addlegendentry{Only Flings}

\addplot+ [lime!50!black, dashed, mark=Mercedes star] table[x=step, y=cov_stretch] {figures/cov-over-step.txt};
\addlegendentry{No Stretching}

\addplot [name path=upper, draw=none] table[x=step, y expr=\thisrow{cov}+\thisrow{cov_err}] {figures/cov-over-step.txt};
\addplot [name path=lower, draw=none] table[x=step, y expr=\thisrow{cov}-\thisrow{cov_err}] {figures/cov-over-step.txt};
\addplot [fill=teal, error] fill between [of=upper and lower];

\addplot [name path=upper_max, draw=none] table[x=step, y expr=\thisrow{cov_max}+\thisrow{cov_max_err}] {figures/cov-over-step.txt};
\addplot [name path=lower_max, draw=none] table[x=step, y expr=\thisrow{cov_max}-\thisrow{cov_max_err}] {figures/cov-over-step.txt};
\addplot [fill=orange, error] fill between [of=upper_max and lower_max];

\addplot [name path=upper_flingbot, draw=none] table[x=step, y expr=\thisrow{cov_flingbot}+\thisrow{cov_flingbot_err}] {figures/cov-over-step.txt};
\addplot [name path=lower_flingbot, draw=none] table[x=step, y expr=\thisrow{cov_flingbot}-\thisrow{cov_flingbot_err}] {figures/cov-over-step.txt};
\addplot [fill=violet, error] fill between [of=upper_flingbot and lower_flingbot];

\addplot [name path=upper_fling, draw=none] table[x=step, y expr=\thisrow{cov_fling}+\thisrow{cov_fling_err}] {figures/cov-over-step.txt};
\addplot [name path=lower_fling, draw=none] table[x=step, y expr=\thisrow{cov_fling}-\thisrow{cov_fling_err}] {figures/cov-over-step.txt};
\addplot [fill=gray, error] fill between [of=upper_fling and lower_fling];

\addplot [name path=upper_stretch, draw=none] table[x=step, y expr=\thisrow{cov_stretch}+\thisrow{cov_stretch_err}] {figures/cov-over-step.txt};
\addplot [name path=lower_stretch, draw=none] table[x=step, y expr=\thisrow{cov_stretch}-\thisrow{cov_stretch_err}] {figures/cov-over-step.txt};
\addplot [fill=lime!50!black, error] fill between [of=upper_stretch and lower_stretch];

\end{axis}

\begin{axis}[
    xlabel={Action Step $t$},
    ylabel={\small Sufficiently Smoothed Rate},
    height=150,
    width=170,
    xmax=9,
    grid=both,
    grid style={dashed, draw=gray},
    major grid style={line width=.2pt,draw=gray!50},
    at={(6cm, 0)},
    % legend pos=south east
]

% \draw[thick, dashed, teal] (3.0, -1) -- ++(0, 3);
% \draw[thick, dashed, orange] (4.9, -1) -- ++(0, 3);
% \draw[thick, dashed, black] (5.3, -1) -- ++(0, 3);

\addplot+ [teal, mark=o] table[x=step,y=hist_cum, teal] {figures/done-over-step.txt};
% \addlegendentry{BiMaMa-Net}

\addplot+ [orange, mark=triangle] table[x=step,y=hist_max_cum] {figures/done-over-step.txt};
% \addlegendentry{Max Value Map}

\addplot+ [black, dashed, mark=star] table[x=step,y=hist_fling_cum] {figures/done-over-step.txt};
% \addlegendentry{Only Flings}

\addplot+ [black, violet, mark=square] table[x=step,y=hist_flingbot_cum] {figures/done-over-step.txt};
% \addlegendentry{Flingbot}

\addplot+ [lime!50!black, dashed, mark=Mercedes star] table[x=step,y=hist_no_cum] {figures/done-over-step.txt};
% \addlegendentry{No Stretching}

\end{axis}

\begin{axis}[
    xlabel={Action Step $t$},
    ylabel={Coverage},
    height=150,
    width=170,
    xmax=9,
    grid=both,
    grid style={dashed, draw=gray},
    major grid style={line width=.2pt,draw=gray!50},
    at={(12cm, 0)},
    legend style={
        at={(0.5, -0.29)},
        anchor=north,
        legend columns=1
    }
]
\addplot+ [teal, mark=o] table[x=step, y=trained, teal] {figures/generalization-cov-over-step.txt};
\addlegendentry{T-shirt (trained)}

\addplot+ [olive, mark=diamond] table[x=step, y=towels] {figures/generalization-cov-over-step.txt};
\addlegendentry{Towels (unseen)}

\addplot+ [magenta, mark=pentagon] table[x=step, y=shirts] 
{figures/generalization-cov-over-step.txt};
\addlegendentry{T-shirts (unseen)}

\addplot [name path=upper, draw=none] table[x=step, y expr=\thisrow{trained}+\thisrow{trained_err}] {figures/generalization-cov-over-step.txt};
\addplot [name path=lower, draw=none] table[x=step, y expr=\thisrow{trained}-\thisrow{trained_err}] {figures/generalization-cov-over-step.txt};
\addplot [fill=teal, error] fill between [of=upper and lower];

\addplot [name path=upper_max, draw=none] table[x=step, y expr=\thisrow{towels}+\thisrow{towels_err}] {figures/generalization-cov-over-step.txt};
\addplot [name path=lower_max, draw=none] table[x=step, y expr=\thisrow{towels}-\thisrow{towels_err}] {figures/generalization-cov-over-step.txt};
\addplot [fill=olive, error] fill between [of=upper_max and lower_max];

\addplot [name path=upper_max, draw=none] table[x=step, y expr=\thisrow{shirts}+\thisrow{shirts_err}] {figures/generalization-cov-over-step.txt};
\addplot [name path=lower_max, draw=none] table[x=step, y expr=\thisrow{shirts}-\thisrow{shirts_err}] {figures/generalization-cov-over-step.txt};
\addplot [fill=magenta, error] fill between [of=upper_max and lower_max];

\end{axis}
\end{tikzpicture}
    \caption{\textbf{Garment smoothing} until it is \done{}. We compare the normalized coverage (left) and prediction of the learned \emph{\done{}} classifier (center) over the number of action steps with different baseline methods. The system is able to generalize to unseen garments of different color, patterns, and material (right).}
    \label{fig:smoothing_over_action}
    \vspace{-20pt}
\end{figure*}

We compare three approaches for folding: \textit{instruction-based folding}, which can be adapted to different garments and different folding techniques, \textit{''2-second`` fold}, a known heuristic for surprisingly fast t-shirt folding, and \textit{fling-to-fold} (F2F), a novel technique that can increase the number of folds-per-hour (FPH) by leveraging prior knowledge about the t-shirt's dimensions.

\begin{description}[leftmargin=4mm]
\item[Instruction-based Folding:] As shown in Fig.~\ref{fig:folding} (left), given a mask of a smoothed garment, the robot iteratively folds the garment along user-specified \emph{folding lines}. These allow to define the goal configuration of a smooth garment precisely without using high-dimensional visual goal representations~\cite{weng2022fabricflownet, hoque2020visuospatial}. A complete user instruction includes: (1) A binary mask called template and (2) a list of folding lines relative to the template (Fig.~\ref{fig:folding} left). The folding direction is defined with respect to the line according to the right-hand-rule. 

To execute the folding lines, a particle-swarm optimizer computes an affine transformation by registering the template with the current image. Afterwards, \sysfull{} calculates corresponding poses for a bimanual fold action: Let $\vec{p}_{ent}$ be the first and $\vec{p}_{exit}$ be the second intersection point of the line and the mask, where the line enters and exits the mask respectively. This splits the mask into a \emph{base} and a \emph{fold-on-top} part. On the contour of the latter, the algorithm finds two pick points $\vec{p}_1$ and $\vec{p}_2$ so that the area of the four-sided polygon $(\vec{p}_{ent}, \vec{p}_{1}, \vec{p}_{2}, \vec{p}_{exit})$ is maximized. We use the normal at the pick point for the gripper orientation $\theta$. The place poses are calculated by mirroring the pick poses at the folding line.

\item[``2-Second'' Fold:] For specific garments such as t-shirts, there exist heuristics for efficient folding. Given a smooth configuration, the ''2-second`` fold follows a set of steps that requires using two arms simultaneously (Fig.~\ref{fig:folding}), and is therefore well suited for a bimanual robot \cite{2sfold}.

\item[Fling-to-fold (F2F):] We observe that (1) a fling action while grasping a sleeve and the non-diagonal bottom corner is especially effective and (2) the first fold action grasps the same points. We conclude that these two steps can be merged to reduce imaging and motion time. We implement F2F by adding a learned primitive to \algname{} that computes these pick points if visible. The primitive's motion is implemented by combining a fling with a consecutive fold action (Fig.~\ref{fig:folding}). To ensure that the t-shirt is folded correctly, prior knowledge about the t-shirt's dimension is required to adapt the height of each arm prior to the fold.

\end{description}
\section{Experiments}
\label{sec:experiments}

We experimentally evaluate the garment smoothing and folding performance of \sysfull{} on a known t-shirt, as well as on two garments unseen during training.

\subsection{Experimental Setup}

We perform experiments on a physical ABB YuMi robot with parallel-jaw grippers. The gripper's fingertips are extended by small 3D printed teeth to improve grasping. A thin sponge mattress is placed on the workspace to allow the grippers to reach below the garment without colliding. A Photoneo PhoXi captures overhead grayscale and depth images of the workspace, generating observations $o_t \in \mathbb{R}^{256 \times 192 \times 2}$. As the garment is frequently outside the camera's field of view, a 1080P GESMATEK RGB webcam is mounted above the workspace and used for coverage calculation. Computing is done on a system using an Intel i7-6850K CPU, 32GB RAM, and a NVIDIA GeForce RTX 2080 Ti.

We first perform data collection, and train~\ref{sec:training} on a single t-shirt. Initially, \num{600} scenes of random garment configuration were recorded and manually annotated in \SI{1}{h}. After training a first \ac{NN}, the robot collected \num{2200} self-supervised actions in \SI{16}{h}. To include data of less frequently observed actions, we copied and re-annotated \num{1500} actions in \SI{3}{h}, resulting in a dataset of \num{4300} actions in total. We used a single t-shirt shown in Fig.~\ref{fig:fig_one} throughout the training.
We further perform data augmentation, including random translations, rotations, flips, resizes, brightness and contrast changes. We use $N=20$ gripper orientations equally distributed over $[0, 2\pi)$ to implement the \algname{} decoder as described in Sec.~\ref{sec:bimama}. For training, we manually tune $N_s=50$, $r_h=0.8$ and $r_c=0.3$ (see Sec.~\ref{sec:training}).
\newcolumntype{Y}{>{\centering\arraybackslash}X}

\begin{table*}[t]
    \vspace{1mm}
    \centering
    \caption{\textbf{End-to-end folding} for different \ac{NN} architectures, folding approaches, and garments, averaged over 15 trials per experiment. The durations are averaged over successful folds, while the cycle time and FPH are averaged over both successful and unsuccessful folds.}
% IMPORTANT: 2s fold is not end-to-end but folding only:
\begin{tabularx}{\linewidth}{cccYYYYY}
    \toprule
    \textbf{Method} & \textbf{Folding Approach} & \textbf{Garment} & \textbf{Smoothing Actions} & \textbf{Duration} [\si{s}] & \textbf{Fold Success} & \textbf{Cycle Time} [\si{s}] & \textbf{Folds Per Hour (FPH)} \\
    \midrule
    Max Value Map & Instruction & T-shirt & \num{5.1 \pm 0.5} & \num{133.9 \pm 7.4} & \SI{80}{\%} & \num{167.4 \pm 9.2} & \num{21.5 \pm 1.2} \\
    \hdashline
    \multirow{3}{*}{\algname{}} & Instruction & \multirow{3}{*}{T-shirt} & \num{3.0 \pm 0.4} & \num{108.7 \pm 7.3} & \textbf{\color{green!30!black!70} \SI{93}{\%}} & \num{116.9 \pm 7.9} & \num{30.8 \pm 2.1} \\
    & ''2-Second`` Fold & & \num{3.0 \pm 0.4} & \num{97.3 \pm 4.8} & \SI{53}{\%} & \num{182.4 \pm 5.4} & \num{19.7 \pm 0.6} \\
    & Fling-to-fold & & \textbf{\color{green!30!black!70} \num{1.8 \pm 0.2}} & \textbf{\color{green!30!black!70} \num{81.7 \pm 4.3}} & \textbf{\color{green!30!black!70} \SI{93}{\%}} & \textbf{\color{green!30!black!70} \num{87.9 \pm 4.7}} & \textbf{\color{green!30!black!70} \num{40.9 \pm 2.2}} \\
    \midrule
    \multicolumn{8}{c}{\textbf{Garments Unseen During Training}} \\
    \midrule
    \multirow{2}{*}{\algname{}} & \multirow{2}{*}{Instruction} & Towel & \num{1.7 \pm 0.2} & \num{59.2 \pm 3.8} & \SI{87}{\%} & \num{68.1 \pm 4.4} & \num{52.9 \pm 3.4} \\
    & & T-shirt & \num{4.8 \pm 0.4} & \num{141.1 \pm 8.7} & \SI{80}{\%} & \num{176.3 \pm 10.9} & \num{20.4 \pm 1.3} \\
    \bottomrule
    \vspace{-20pt}
\end{tabularx}
    \label{tab:main-results}
\end{table*}

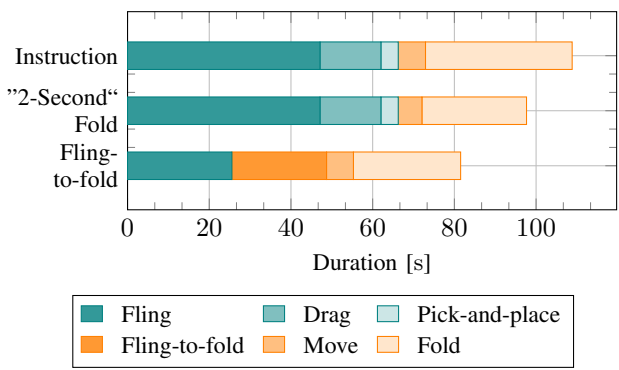
\begin{figure}[b]
    \centering
\begin{tikzpicture}
    \begin{axis}[
        xbar stacked,
        ytick=data,
        yticklabels={Fling-to-fold, ''2-Second`` Fold, Instruction},
        width=230,
        height=120,
        bar width=0.5,
        xmin=0,
        enlarge y limits=0.4,
        grid,
        minor tick num=2,
        yticklabel style={text width=1.6cm, align=right, font=\small},
        xlabel={\small Duration [\si{s}]},
        legend style={
            at={(0.4, -0.43)},
            anchor=north,
            legend columns=3,
            column sep=5pt,
            font=\small
        },
        legend cell align={left},
    ]
\addplot[teal, fill=teal!80] coordinates
% Fling
{(25.54,0) (47.11,1) (47.11,2)};
\addplot[teal, fill=teal!50] coordinates
% Drag
{(0.00,0) (14.91,1) (14.91,2)};
\addplot[teal, fill=teal!20] coordinates
% Pick-and-place (single)
{(0.00,0) (4.24,1) (4.24,2)};
\addplot[orange, fill=orange!80] coordinates
% Fling-to-fold
{(23.17,0) (0.00,1) (0.00,2)};
\addplot[orange, fill=orange!50] coordinates
% Move
{(6.58,0) (5.80,1) (6.67,2)};
\addplot[orange, fill=orange!20] coordinates
% Pick-and-place (dual)
{(26.24,0) (25.60,1) (35.90,2)};

\legend{Fling, Drag, Pick-and-place, Fling-to-fold, Move, Fold}
\end{axis}
\end{tikzpicture}
    \caption{\textbf{Timings} for calculating and executing the action primitive types depending on the folding approach. Instruction-based and ''2-Second`` fold share the same smoothing actions (blue), but differ in folding (orange). By introducing a combined Fling-to-fold primitive, a smoothed state is not required before folding. However, the ''2-Second`` fold is suited for manipulating a t-shirt only, and Fling-to-fold assumes prior knowledge of the garment.}
    \label{fig:fold-timing}
\end{figure}

We design a set of garment smoothing and folding experiments to evaluate \sysfull{}. Initial garment configurations are generated by environment resets as described in \ref{sec:training}. Each experiment is averaged over 15 trials. We ignore experiments that terminate early due to a motion planning error, as this is not the focus of this paper. A trial is considered unsuccessful if the garment was not successfully folded according to the majority vote of three reviewers or the number of actions exceeded a maximal horizon of $H=10$. We define a grasp success if the gripper holds the garment \emph{after} an executed action. For known garments, \algname{} achieves a grasp success rate of over \SI{96}{\%}.

\subsection{\done{}}

We evaluate garment smoothing using two metrics: The garment coverage, computed from an overhead image, and a binary \done{} value, predicted using the \done{} classifier. We compare \algname{} to two baseline (1) \textit{Max Value Map}, a variant of \algname{} that computes the pick points directly from the value maps $Q_{unc}$ by computing the maximum over the map to find two pick points without using correspondence descriptors, (2) \textit{Only Flings}, a variant restricted to fling actions only, and (3) \textit{Flingbot}, the pre-trained method from~\cite{ha2022flingbot} (see Fig.~\ref{fig:smoothing_over_action}). Results suggest that \algname{} is able to smooth a known t-shirt to a \done{} configuration in $\sim3$ fewer steps compared to baselines requiring $\sim5$. Although the increase of coverage is similar to Only Flings, the latter reaches a \done{} configuration later or even fails to do so, confirming the need of additional action primitives to fully smooth a garment.

We note that the FlingBot baseline fails to reach an \SI{80}{\%} coverage as reported in~\cite{ha2022flingbot}, as we observe frequent grasp failures presumably due to differences in the physical setting. We ablate the stretching motion before a fling and observe that stretching leads to higher coverage.  

\subsection{Folds per Hour}

Table~\ref{tab:main-results} shows results of end-to-end garments folding experiments. \algname{} manages to (1) successfully fold garments in over \SI{90}{\%} of the trials on known garments and (2) \SI{30}{\%} faster than the Max Value Map baseline using the Instruction-based folding approach. The ''2-second`` fold achieves an additional speedup of \SI{10.4}{\%} when executed successfully, however we find that it is sensitive to t-shirt's orientation in a \done{} configuration and suffers from a low fold success rate. With prior information on the t-shirt's dimensions, F2F uses \SI{40}{\%} less smoothing actions and imaging time. As a result, it achieves a speedup of over \SI{25}{\%} compared to the instruction-based approach, leading to \num{40.9} folds per hour on average. Calculating an action using \algname{} takes \SI{126.0 \pm 0.9}{ms} on our hardware.

\subsection{Generalization to Unseen Garments}
\label{sec:generalization}

We explore how \sysfull{}, trained on a single t-shirt, can generalize to garments unseen during training. In these experiments we use (1) a t-shirt with a different color and stiffness and (2) a rectangular towel with a different color compared to the original t-shirt. We evaluate \sysfull{} on unseen garments using instruction-based folding, as this is the only approach that easily adapts to general garments. We run the same experiments on the unseen t-shirt with no changes to the \algname{} model or the folding template. In contrast, when we run the towel experiments we observe that the system fails to classify a \done{} configuration, as the object's shape is different from that \algname{} was trained on. To address this, we add \num{20} \done{} towel images to the dataset and re-train \algname{}. Table~\ref{tab:main-results} suggests that \sysfull{} can generalize to garments with different color, stiffness and shape.

\subsection{System Limitations}

Grasp failures, especially during a fling motion, can decrease the garment's coverage dramatically. We find that most grasp failures happen due to losing the grip during the stretching motion prior to a fling action. This limitation can be mitigated using improved force feedback or by adding visual feedback. We observe a frequent failure case during top-down grasps while executing the first step of the ''2-second`` fold. These grasps may require different gripper jaws that are better suited for top-down grasps. 

As common with data-driven methods, \sysfull{} can generalize to \emph{similar} unseen garments. For example, textile patterns may be more challenging to detect and classify correctly. This limitation can be addressed through additional data augmentation. Generalization to different garment shapes may also be limited, and can be addressed by adding examples of \done{} configurations to the dataset, as described in Sec.~\ref{sec:generalization} for the towel example.
\section{Conclusion and Discussion}
\label{sec:conclusion}

We presented \sysfull{}, a bimanual robotic system for efficient folding of garments from arbitrary initial configurations. At its core, a novel \algname{} architecture predicts two conditioned poses to parameterize a set of manipulation primitives. After learning from \num{4300} human-annotated or self-supervised actions, the robot is able to fold garments in under \SI{120}{s} on average with a success rate of \SI{93}{\%}.

While prior works e.g.\ by Maitin-Shepard et al.~\cite{maitin2010cloth} or Doumanoglou et al.~\cite{doumanoglou2016folding} achieved high success rate for end-to-end cloth folding, cycle times for a single fold were on the order of $3-6$ Folds Per Hour (FPH), whereas \sysfull{} achieves \SIrange{30}{40}{FPH}.
Similar to Ha et al.~\cite{ha2022flingbot}, the fling primitive can unfold the garment in a few actions. In contrast however, we introduce additional action primitives that enable the robot to reach a sufficiently smoothed configuration. In future work, we will explore methods that can learn to manipulate a novel garment given a few demonstrations.

\section*{ACKNOWLEDGMENT}

\small{This research was performed at the AUTOLAB at UC Berkeley in affiliation with the Berkeley AI Research (BAIR) Lab, Berkeley Deep Drive (BDD), the Real-Time Intelligent Secure Execution (RISE) Lab, and the CITRIS “People and Robots” (CPAR) Initiative. We thank Max Cao and Huy Ha for their helpful feedback.}

\balance
\bibliographystyle{IEEEtran}
\bibliography{./root}

\end{document}